\documentclass[10pt,twocolumn,letterpaper]{article}
\pdfoutput=1
\usepackage{cvpr}
\usepackage{times}
\usepackage{epsfig}
\usepackage{graphicx}
\usepackage{amsmath}
\usepackage{amssymb}

\usepackage{enumitem}
\usepackage{graphicx}
\usepackage{amsmath,amssymb} 
\usepackage{subcaption}
\usepackage{array}
\newcolumntype{P}[1]{>{\centering\arraybackslash}p{#1}}
\newcolumntype{M}[1]{ >{\centering\arraybackslash} m{#1} }
\newcommand{\rowgroup}[1]{\hspace{-0em}#1}

\usepackage{relsize}
\usepackage{verbatim}
\usepackage{multirow}
\usepackage[title]{appendix}

\newcommand{\id}{\mathrm{id}}

\newif\ifvae
\vaefalse

\usepackage[pagebackref=true,breaklinks=true,letterpaper=true,colorlinks,bookmarks=false]{hyperref}

\cvprfinalcopy 

\def\cvprPaperID{6477} 

\ifcvprfinal\fi
\begin{document}

\title{Data augmentation using learned transformations\\ for one-shot medical image segmentation}

\author{
Amy Zhao\\
MIT\\
{\tt\small xamyzhao@mit.edu}
\and
\hspace{2cm}Guha Balakrishnan\\
\hspace{2cm}MIT\\
\hspace{2cm}{\tt\small balakg@mit.edu} 
\and
\hspace{2cm}Fr\'edo Durand\\
\hspace{2cm}MIT\\
\hspace{2cm}{\tt\small fredo@mit.edu}
\and
John V. Guttag\\
MIT\\
{\tt\small guttag@mit.edu}
\and
Adrian V. Dalca\\
MIT, MGH\\
{\tt\small adalca@mit.edu}\\
}
\maketitle

\begin{abstract}
\vspace{-5pt}
Image segmentation is an important task in many medical applications. Methods based on convolutional neural networks attain state-of-the-art accuracy; however, they typically rely on supervised training with large labeled datasets. Labeling medical images requires significant expertise and time, and typical hand-tuned approaches for data augmentation fail to capture the complex variations in such images. 

We present an automated data augmentation method for synthesizing labeled medical images. We demonstrate our method on the task of segmenting magnetic resonance imaging (MRI) brain scans. Our method requires only a single segmented scan, and leverages other unlabeled scans in a semi-supervised approach. We learn a model of transformations from the images, and use the model along with the labeled example to synthesize additional labeled examples. Each transformation is comprised of a spatial deformation field and an intensity change, enabling the synthesis of complex effects such as variations in anatomy and image acquisition procedures. We show that training a supervised segmenter with these new examples provides significant improvements over state-of-the-art methods for one-shot biomedical image segmentation. 
\end{abstract}

\section{Introduction}

\begin{figure}[t!]
\centering
\includegraphics[width=\linewidth]{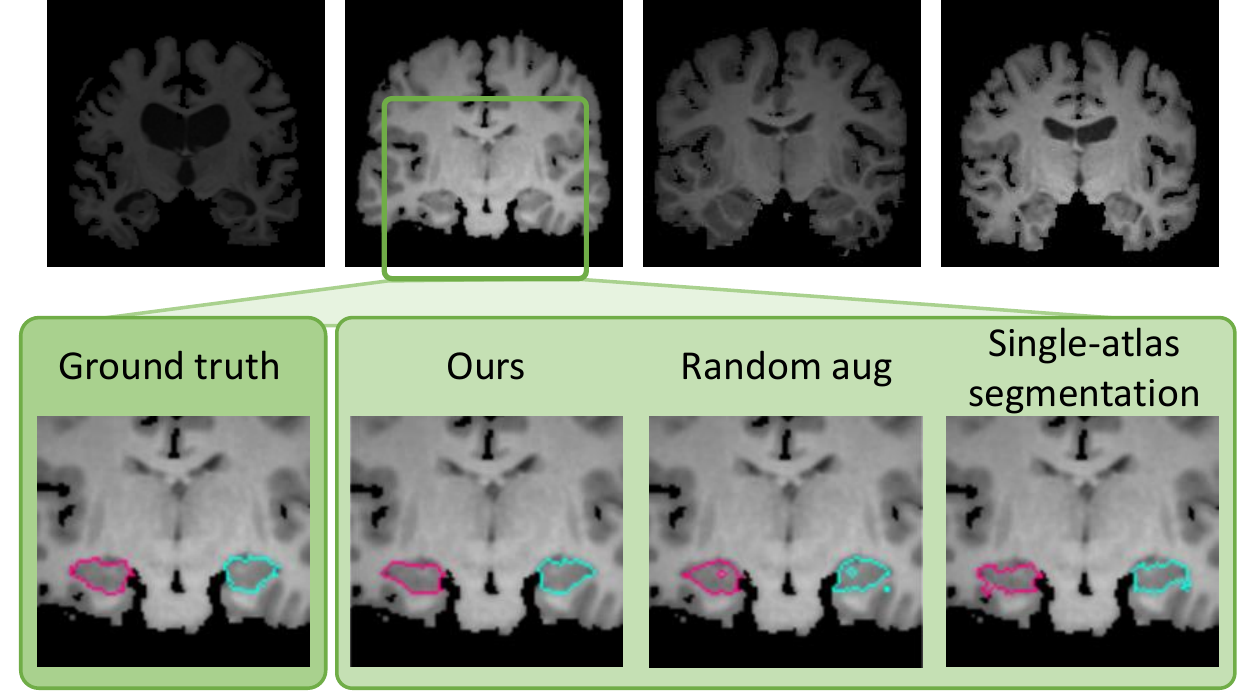}
\vspace{-16pt}\caption{Biomedical images often vary widely in anatomy, contrast and  texture (top row). Our method enables more accurate segmentation of anatomical structures compared to other one-shot segmentation methods (bottom row).}
\label{fig:teaser}
\end{figure}

Semantic image segmentation is crucial to many biomedical imaging applications, such as performing population analyses, diagnosing disease, and planning treatments. When enough labeled data is available, supervised deep learning-based segmentation methods produce state-of-the-art results. However, obtaining manual segmentation labels for medical images requires considerable expertise and time. In most clinical image datasets, there are very few manually labeled images. The problem of limited labeled data is exacerbated by differences in image acquisition procedures across machines and institutions, which can produce wide variations in resolution, image noise, and tissue appearance \cite{leung2010robust}. 

To overcome these challenges, many supervised biomedical segmentation methods focus on hand-engineered preprocessing steps and architectures \cite{moeskops2016automatic,pereira2016brain}. It is also common to use hand-tuned data augmentation to increase the number of training examples ~\cite{akkus2017deep,oliveira2017augmenting,pereira2016brain,ronneberger2015u,roth2015deeporgan}. Data augmentation functions such as random image rotations or random nonlinear deformations are easy to implement, and are effective at improving segmentation accuracy in some settings \cite{oliveira2017augmenting,pereira2016brain,ronneberger2015u,roth2015deeporgan}. However, these functions have limited ability to emulate real variations \cite{eaton2018improving}, and can be highly sensitive to the choice of parameters~\cite{dosovitskiy2016discriminative}.

We address the challenges of limited labeled data by learning to synthesize diverse and realistic labeled examples. Our novel, automated approach to data augmentation leverages unlabeled images. Using learning-based registration methods, we model the set of spatial and appearance transformations between images in the dataset. These models capture the anatomical and imaging diversity in the unlabeled images. We synthesize new examples by sampling transformations and applying them to a single labeled example. 

We demonstrate the utility of our method on the task of one-shot segmentation of brain magnetic resonance imaging (MRI) scans. We use our method to synthesize new labeled training examples, enabling the training of a supervised segmentation network. This strategy outperforms state-of-the art one-shot biomedical segmentation approaches, including single-atlas segmentation and supervised segmentation with hand-tuned data augmentation. 

\begin{figure*}[t]
\centering
{\includegraphics[width=\linewidth]
{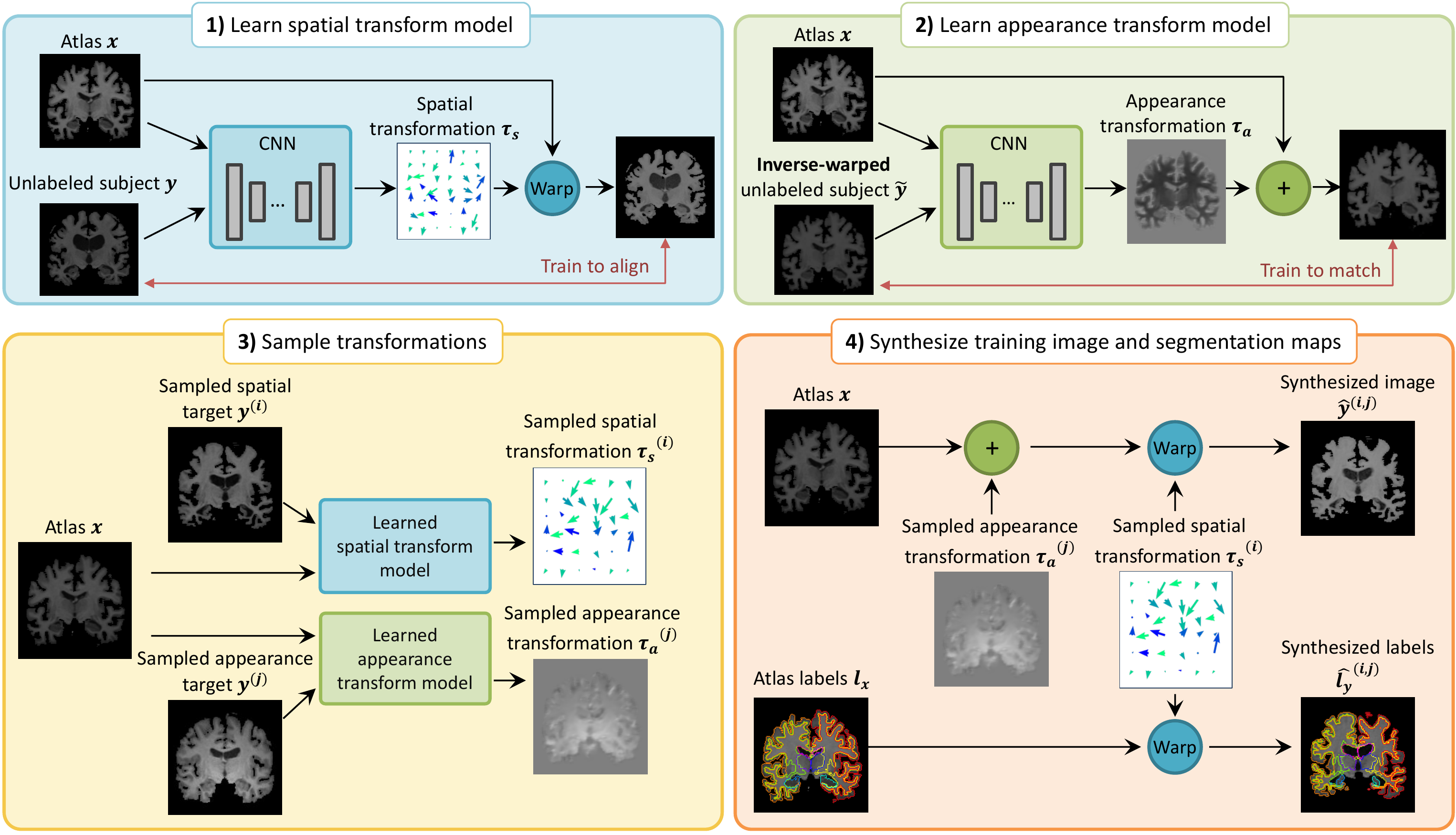}}
\caption{An overview of the proposed method. We learn independent spatial and appearance transform models to capture the variations in our image dataset. We then use these models to synthesize a dataset of labeled examples. This synthesized dataset is used to train a supervised segmentation network.}
\label{fig:overview}
\end{figure*}

\section{Related work}
\subsection{Medical image segmentation}
We focus on the segmentation of brain MR images, which is challenging for several reasons. Firstly, human brains exhibit substantial anatomical variations \cite{frost2012measuring,rademacher2001variability,van2007surface}. Secondly, MR image intensity can vary as a result of subject-specific noise, scanner protocol and quality, and other imaging parameters \cite{leung2010robust}. This means that a tissue class can appear with different intensities across images -- even images of the same MRI modality. 

Many existing segmentation methods rely on scan pre-processing to mitigate these intensity-related challenges. Pre-processing methods can be costly to run, and developing techniques for realistic datasets is an active area of research \cite{glocker2018nonparametric,sridharan2013quantification}. Our augmentation method tackles these intensity-related challenges from another angle: rather than removing intensity variations, it enables a segmentation method to be robust to the natural variations in MRI scans.

A large body of classical segmentation methods use \textit{atlas-based} or \textit{atlas-guided} segmentation, in which a labeled reference volume, or \textit{atlas}, is aligned to a target volume using a deformation model, and the labels are propagated using the same deformation \cite{baillard2001segmentation,ciofolo2009atlas,dawant1999automatic,hellier2004hierarchical}.  When multiple atlases are available, they are each aligned to a target volume, and the warped atlas labels are fused \cite{iglesias2015multi,klein2005mindboggle,sabuncu2010generative,wang2013multi}.  In atlas-based approaches, anatomical variations between subjects are captured by a deformation model, and the challenges of intensity variations are mitigated using pre-processed scans, or intensity-robust metrics such as normalized cross-correlation. However, ambiguities in tissue appearances (\textit{e.g.}, indistinct tissue boundaries, image noise) can still lead to inaccurate registration and segmentations. We address this limitation by training a segmentation model on diverse realistic examples, making the segmenter more robust to such ambiguities. We focus on having a single atlas, and demonstrate that our strategy outperforms atlas-based segmentation. If more than one segmented example is available, our method can leverage them.

Supervised learning approaches to biomedical segmentation have gained popularity in recent years. To mitigate the need for large labeled training datasets, these methods often use data augmentation along with hand-engineered pre-processing steps and architectures \cite{akkus2017deep,kamnitsas2017efficient,moeskops2016automatic,pereira2016brain,ronneberger2015u,roth2015deeporgan,zhang2015deep}. 

Semi-supervised and unsupervised approaches have also been proposed to combat the challenges of small training datasets. These methods do not require paired image and segmentation data. Rather, they leverage collections of segmentations to build anatomical priors~\cite{dalca2018anatomical}, to train an adversarial network \cite{joyce2018deep}, or to train a novel semantic constraint \cite{ganaye2018semi}. In practice, collections of images are more readily available than segmentations. Rather than rely on segmentations, our method leverages a set of unlabeled images.


\subsection{Spatial and appearance transform models}
Models of shape and appearance have been used in a variety of image analyses. Parametric spatial transform models have been used to align and classify handwritten digits~\cite{hauberg2016dreaming,learned2006data,miller2000learning}. In medical image registration, a spatial deformation model is used to establish semantic correspondences between images. This mature field spans optimization-based methods \cite{ashburner2000,bajcsy1989,rueckert1999,shen2002}, and recent learning-based methods \cite{ balakrishnan2018unsupervised,balakrishnan2019voxelmorph,dalca2018unsupervised,krebs2017,rohe2017,sokooti2017,yang2017}.  We leverage VoxelMorph \cite{balakrishnan2018unsupervised,balakrishnan2019voxelmorph}, a recent unsupervised learning-based method, to learn spatial transformations.

Many medical image registration methods focus on intensity-normalized images or intensity-independent objective functions, and do not explicitly account for variations in image intensity. For unnormalized images, models of intensity transforms have used to remove bias field effects from MRI \cite{learned2006data,wells1996adaptive}. Spatial and appearance transform models have been used together to register objects that differ in shape as well as texture. Many works build upon the framework of Morphable Models \cite{jones1998multidimensional} or Active Appearance Models (AAMs)~\cite{cootes1999unified,cootes2001active}, in which statistical models of shape and texture are constructed. AAMs have been used to localize anatomical landmarks \cite{cootes2001statistical,potesil2015personalized} and perform segmentation \cite{mitchell20023,patenaude2011bayesian,vincent2012fully}. We build upon these concepts by using convolutional neural networks to learn models of unconstrained spatial and intensity transformations. Rather than learning transform models for the end goal of registration or segmentation, we sample from these models to synthesize new training examples. As we show in our experiments, augmenting a segmenter's training set in this way can produce more robust segmentations than performing segmentation using the transform models directly.

\subsection{Few-shot segmentation of natural images}
Few-shot segmentation is a challenging task in semantic segmentation and video object segmentation. Existing approaches focus mainly on natural images. Methods for few-shot semantic segmentation incorporate information from prototypical examples of the classes to be segmented \cite{dong2018few,shaban2017one}. Few-shot video segmentation is frequently implemented by aligning objects in each frame to a labeled reference frame \cite{jain2017fusionseg, tsai2016video}. Other approaches leverage large labeled datasets of supplementary information such as object appearances \cite{caelles2017one}, or incorporate additional information such as human input~\cite{rakelly2018few}. Medical images present different challenges from natural images; for instance, the visual differences between tissue classes are very subtle compared to the differences between objects in natural images.

\subsection{Data augmentation}
In image-based supervised learning tasks, data augmentation is commonly performed using simple parameterized transformations such as rotation and scaling. For medical images, random smooth flow fields have been used to simulate anatomical variations \cite{milletari2016v,ronneberger2015u,roth2015anatomy}. These parameterized transformations can reduce overfitting and improve test performance \cite{huang2016densely,krizhevsky2012imagenet,milletari2016v,ronneberger2015u,roth2015anatomy}. However, the performance gains imparted by these transforms vary with the selection of transformation functions and parameter settings~\cite{dosovitskiy2016discriminative}. 

Recent works have proposed learning data augmentation transformations from data. Hauberg \textit{et al.} \cite{hauberg2016dreaming} focus on data augmentation for classifying MNIST digits. They learn digit-specific spatial transformations, and sample training images and transformations to create new examples aimed at improving classification performance. We learn an appearance model in addition to a spatial model, and we focus on the problem of MRI segmentation. 
Other recent works focus on learning combinations of simple transformation functions (\textit{e.g.}, rotation and contrast enhancement) to perform data augmentation for natural images \cite{cubuk2018autoaugment,ratner2017learning}. Cubuk \textit{et al.} \cite{cubuk2018autoaugment} use a search algorithm to find augmentation policies that maximize classification accuracy. Ratner \textit{et al.} \cite{ratner2017learning} learn to create combinations of transformations by training a generative adversarial network on user input. These simple transformations are insufficient for capturing many of the subtle variations in MRI data.

\section{Method}
\begin{figure*}[t]
    \centering
        \includegraphics[width=\linewidth]{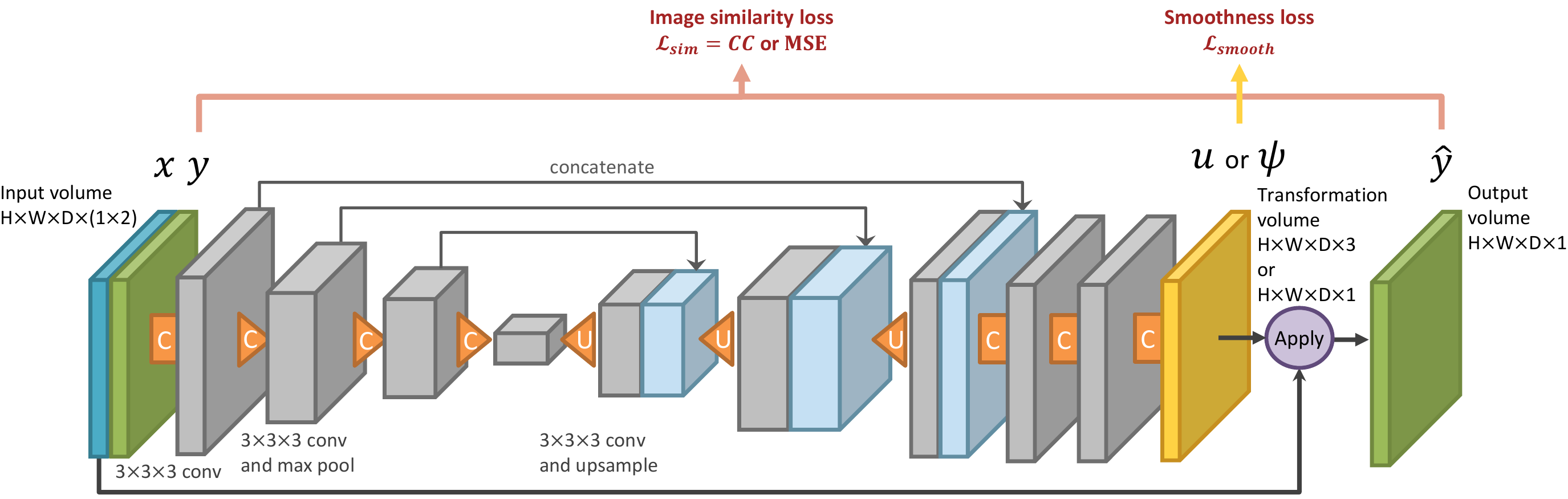}
    \caption{We use a convolutional neural network based on the U-Net architecture \cite{ronneberger2015u} to learn each transform model. The application of the transformation is a spatial warp for the spatial model, and a voxel-wise addition for the appearance model. Each convolution uses $3\times3\times3$ kernels, and is followed by a LeakyReLU activation layer. The encoder uses max pooling layers to reduce spatial resolution, while the decoder uses upsampling layers. }\label{fig:network}
\end{figure*}

We propose to improve one-shot biomedical image segmentation by synthesizing realistic training examples in a semi-supervised learning framework. 

Let $\{y^{(i)}\}$ be a set of biomedical image volumes, and let the pair $(x, l_x)$ represent a labeled reference volume, or \textit{atlas}, and its corresponding segmentation map. In brain MRI segmentation, each $x$ and $y$ is a grayscale 3D volume. We focus on the challenging case where only one labeled atlas is available, since it is often difficult in practice to obtain many segmented volumes. Our method can be easily extended to leverage additional segmented volumes. 

To perform data augmentation, we apply transformations $\tau^{(k)}$ to the labeled atlas $x$. We first learn separate spatial and appearance transform models to capture the distribution of anatomical and appearance differences between the labeled atlas and each unlabeled volume. Using the two learned models, we synthesize labeled volumes $\{(\hat y^{(k)}, \hat l_y^{(k)})\}$ by applying a spatial transformation and an appearance transformation to the atlas volume, and by warping the atlas label maps using the spatial transformation. Compared to single-atlas segmentation, which suffers from uncertainty or errors in the spatial transform model, we use the same spatial transformation to synthesize the volume \textit{and} label map, ensuring that the newly synthesized volume is correctly labeled. These synthetic examples form a labeled dataset that characterizes the anatomical and appearance variations in the unlabeled dataset. Along with the atlas, this new training set enables us to train a supervised segmentation network. This process is outlined in Fig. \ref{fig:overview}. 

\subsection{Spatial and appearance transform models}\label{sec:transform_models}
We describe the differences between scans using a combination of spatial and intensity transformations. Specifically, we define a transformation $\tau(\cdot)$ from one volume to another as a composition of a spatial transformation $\tau_{s}(\cdot)$ and an intensity or \textit{appearance} transformation $\tau_{a}(\cdot)$, \textit{i.e.}, $\tau(\cdot) = \tau_s(\tau_a(\cdot))$. 


We assume a spatial transformation takes the form of a smooth voxel-wise displacement field $u$. Following the medical registration literature, we define the \textit{deformation} function $\phi = \id + u$, where $\id$ is the identity function. We use $x \circ \phi$ to denote the application of the deformation $\phi$ to $x$. To model the distribution of spatial transformations in our dataset, we compute the deformation that warps atlas $x$ to each volume $y^{(i)}$ using $\phi^{(i)} = g_{\theta_s}(x, y^{(i)})$, where $g_{\theta_s}(\cdot, \cdot)$ is a parametric function that we describe later.  We write approximate inverse deformation of $y^{(i)}$ to $x$ as ${\phi^{-1}}^{(i)} = g_{\theta_s}(y^{(i)},x)$. 

We model the appearance transformation $\tau_a(\cdot)$ as per-voxel addition in the spatial frame of the atlas. We compute this per-voxel volume using the function \mbox{$\psi^{(i)} = h_{\theta_a}(x, y^{(i)} \circ {\phi^{-1}}^{(i)})$}, where $y^{(i)} \circ {\phi^{-1}}^{(i)}$ is a volume that has been registered to the atlas space using our learned spatial model. 
In summary, our spatial and appearance transformations are:
\vspace{-1pt}
\begin{align}
\tau^{(i)}_s(x) &= x \circ \phi^{(i)}, &\phi=g_{\theta_s}(x, y^{(i)}) \label{eq:spatial_model}\\
\tau^{(i)}_a(x) &= x + \psi^{(i)}, &\psi^{(i)} = h_{\theta_a}(x, y^{(i)}\circ {\phi^{-1}}^{(i)}).\label{eq:app_model}
\end{align}


\subsection{Learning}
We aim to capture the distributions of the transformations $\tau_s$ and $\tau_a$ between the atlas and the unlabeled volumes. We estimate the functions $g_{\theta_s}(\cdot, \cdot)$ and $h_{\theta_a}(\cdot, \cdot)$ in Eqs. \eqref{eq:spatial_model} and \eqref{eq:app_model} using separate convolutional neural networks, with each network using the general architecture outlined in Fig.~\ref{fig:network}. Drawing on insights from Morphable Models~\cite{jones1998multidimensional} and Active Appearance Models~\cite{cootes2001active,cootes2001statistical}, we optimize the spatial and appearance models independently. 

For our spatial model, we leverage VoxelMorph~\cite{balakrishnan2018unsupervised,balakrishnan2019voxelmorph,dalca2018unsupervised}, a recent unsupervised learning-based approach with an open-source implementation. VoxelMorph learns to output a smooth displacement vector field that registers one image to another by jointly optimizing an image similarity loss and a displacement field smoothness term. We use a variant of VoxelMorph with normalized cross-correlation as the image similarity loss, enabling the estimation of $g_{\theta_s}(\cdot,\cdot)$ with unnormalized input volumes. 

We use a similar approach to learn the appearance model. Naively, one might define $h_{\theta_a}(\cdot, \cdot)$ from Eq. \eqref{eq:app_model} as a simple per-voxel subtraction of the volumes in the atlas space. While this transformation would perfectly reconstruct the target image, it would include extraneous details when the registration function $\phi^{-1}$ is imperfect, resulting in image details in $x + \psi$ that do not match the anatomical labels. We instead design $h_{\theta_a}(\cdot, \cdot)$ as a neural network that produces a per-voxel intensity change in an anatomically consistent manner. Specifically, we use an image similarity loss as well as a semantically-aware smoothness regularization. Given the network output $\psi^{(i)} = h_{\theta_a}(x, y^{(i)}\circ \phi^{-1})$, we define a smoothness regularization function based on the atlas segmentation map:
\vspace{-2pt}
\begin{align}
\mathcal{L}_{smooth}(c_x,\psi) &= (1-c_x)\nabla \psi,
\end{align}
where $c_x$ is a binary image of anatomical boundaries computed from the atlas segmentation labels $l_x$, and $\nabla$ is the spatial gradient operator. Intuitively, this term discourages dramatic intensity changes within the same anatomical region. 

In the total appearance transform model loss $\mathcal{L}_a$, we use mean squared error for the image similarity loss $\mathcal{L}_{sim}(\hat{y}, y)=||\hat{y} - y||^2$. In our experiments, we found that computing the image similarity loss in the spatial frame of the subject was helpful. We balance the similarity loss with the regularization term  $\mathcal{L}_{smooth}$:
\vspace{-1pt}
\begin{align*}
&\mathcal{L}_{a}(x,y^{(i)}, \phi^{(i)}, {\phi^{-1}}^{(i)},\psi^{(i)}, c_x) \\
&= \mathcal{L}_{sim}\big((x + \psi^{(i)}) \circ \phi^{(i)}, y^{(i)}\big) + \lambda_a \mathcal{L}_{smooth}(c_x,\psi^{(i)}),
\end{align*}
where $\lambda_a$ is a hyperparameter.

\subsection{Synthesizing new examples}\label{sec:methods_sampling}
The models described in Eqs. \eqref{eq:spatial_model} and \eqref{eq:app_model} enable us to sample spatial and appearance transformations $\tau^{(i)}_s, \tau^{(j)}_a$ by sampling target volumes $y^{(i)}, y^{(j)}$ from an unlabeled dataset. Since the spatial and appearance targets can be different subjects, our method can combine the spatial variations of one subject with the intensities of another into a single synthetic volume $\hat{y}$. We create a labeled synthetic example by applying the transformations computed from the target volumes to the labeled atlas: 
\vspace{-1pt}
\begin{align*}
\hat y^{(i,j)}&=\tau^{(i)}_s(\tau^{(j)}_a(x)),\nonumber\\
\hat l^{(i,j)}_y &=\tau^{(i)}_s(l_x).
\end{align*}
This process is visualized in steps 3 and 4 in Fig. \ref{fig:overview}. 
These new labeled training examples are then included in the labeled training set for a supervised segmentation network.

\subsection{Segmentation network}
The newly synthesized examples are useful for improving the performance of a supervised segmentation network. We demonstrate this using a network based on the state-of-the-art architecture described in \cite{roy2017error}. To account for GPU memory constraints, the network is designed to segment one slice at a time. We train the network on random slices from the augmented training set. We select the number of training epochs using early stopping on a validation set. We emphasize that the exact segmentation network architecture is not the focus of this work, since our method can be used in conjunction with any supervised segmentation network.

\subsection{Implementation}
We implemented all models using Keras \cite{chollet2015} and Tensorflow \cite{abadi2016}. The application of a spatial transformation to an image is implemented using a differentiable 3D spatial transformer layer \cite{balakrishnan2018unsupervised}; a similar layer that uses nearest neighbor interpolation is used to transform segmentation maps. For simplicity, we capture the forward and inverse spatial transformations described in Section \ref{sec:transform_models} using two identical neural networks. For the appearance transform model, we use the hyperparameter setting $\lambda_a=0.02$. We train our transform models with a single pair of volumes in each batch, and train the segmentation model with a batch size of $16$ slices. All models are trained with a learning rate of $5e^{-4}$. Our code is available at \texttt{https://github.com/xamyzhao/brainstorm}.
\section{Experiments}\label{sec:results}
We demonstrate that our automatic augmentation method can be used to improve brain MRI segmentation. We focus on one-shot segmentation of unnormalized scans -- a challenging but practical scenario. Intensity normalization methods such as bias field correction \cite{fischl2012,sled1998nonparametric,styner2000parametric} can work poorly in realistic situations (\textit{e.g.,} clinical-quality scans, or scans with stroke~\cite{sridharan2013quantification} or traumatic brain injury). 

\subsection{Data}
We use the publicly available dataset of T1-weighted MRI brain scans described in \cite{balakrishnan2018unsupervised}. The scans are compiled from eight databases: ADNI~\cite{mueller2005ways}, OASIS~\cite{marcus2007open}, ABIDE~\cite{di2014autism}, ADHD200~\cite{milham2012adhd}, MCIC~\cite{gollub2013mcic}, PPMI~\cite{marek2011parkinson}, HABS~\cite{dagley2017harvard}, and Harvard GSP~\cite{holmes2015brain}; the segmentation labels are computed using FreeSurfer~\cite{fischl2012}. As in \cite{balakrishnan2018unsupervised}, we resample the brains to $256\times 256 \times 256$ with 1mm isotropic voxels, and affinely align and crop the images to $160 \times 192 \times 224$. We do not apply any intensity corrections, and we perform skull-stripping by zeroing out voxels with no anatomical label. For evaluation, we use segmentation maps of the $30$ anatomical labels described in \cite{balakrishnan2018unsupervised}.

We focus on the task of segmentation using a single labeled example. We randomly select $101$ brain scans to be available at training time. In practice, the atlas is usually selected to be close to the anatomical average of the population. We select the most similar training example to the anatomical average computed in \cite{balakrishnan2018unsupervised}. This atlas is the single labeled example that is used to train our transform models; the segmentation labels of the other $100$ training brains are not used. We use an additional $50$ scans as a validation set, and an additional $100$ scans as a held-out test set. 

\subsection{Segmentation baselines}\label{sec:baselines}
\begin{itemize}[leftmargin=-2pt]
\item[] \textbf{Single-atlas segmentation \textit{(SAS)}}: We use the same state-of-the-art registration model \cite{balakrishnan2018unsupervised} that we trained for our method's spatial transform model in a single-atlas segmentation framework. We register the atlas to each test volume, and warp the atlas labels using the computed deformation field~\cite{baillard2001segmentation,ciofolo2009atlas,dawant1999automatic,hellier2004hierarchical,klein2005mindboggle}. That is, for each test image $y^{(i)}$, we compute $\phi^{(i)} = g_{\theta_s}(x, y^{(i)})$ and predict labels $\hat l^{(i)}_{y} = l_x \circ \phi^{(i)}$. 

\item[] \textbf{Data augmentation using single-atlas segmentation \textit{(SAS-aug)}}: We use SAS results as labels for the unannotated training brains, which we then include as training examples for supervised segmentation. This adds $100$ new training examples to the segmenter training set. 

\item[] \textbf{Hand-tuned random data augmentation \textit{(rand-aug)}}: Similarly to \cite{milletari2016v,ronneberger2015u,roth2015anatomy}, we create random smooth deformation fields by sampling random vectors on a sparse grid, and then applying bilinear interpolation and spatial blurring. We evaluated several settings for the amplitude and smoothness of the deformation field, including the ones described in \cite{ronneberger2015u}, and selected the settings that resulted in the best segmentation performance on a validation set. We synthesize variations in imaging intensity using a global intensity multiplicative factor sampled uniformly from the range $[0.5, 1.5]$, similarly to \cite{hussain2017differential,kamnitsas2017efficient}. We selected the range to match the intensity variations in the dataset; this is representative of how augmentation parameters are tuned in practice. This augmentation method synthesizes a new randomly transformed brain in each training iteration.

\item[] \textbf{Supervised}: We train a fully-supervised segmentation network that uses ground truth labels for all $101$ examples in our training dataset. Apart from the atlas labels, these labels are not available for any of the other methods. This method serves as an upper bound.\\
\end{itemize}

\subsection{Variants of our method}
\begin{itemize}[leftmargin=-2pt]
\item[] \textbf{Independent sampling (\textit{ours-indep})}: As described in Section \ref{sec:methods_sampling}, we sample spatial and appearance target images independently to compute $\tau^{(i)}_s, \tau^{(j)}_a$. With $100$ unlabeled targets, we obtain $100$ spatial and $100$ appearance transformations, enabling the synthesis of $10,000$ different labeled examples. Due to memory constraints, we synthesize a random labeled example in each training iteration, rather than adding all $10,000$ new examples to the training set.

\item[] \textbf{Coupled sampling (\textit{ours-coupled})}: To highlight the efficacy of our independent transform models, we compare \textit{ours-indep} to a variant of our method where we sample each of the spatial \textit{and} appearance transformations from the same target image. This results in $100$ possible synthetic examples. As in \textit{ours-indep}, we synthesize a random example in each training iteration.

\item[] \textbf{\textit{Ours-indep} + \textit{rand-aug}}: When training the segmenter, we alternate between examples synthesized using \textit{ours-indep}, and examples synthesized using \textit{rand-aug}. The addition of hand-tuned augmentation to our synthetic augmentation could introduce additional variance that is unseen even in the unlabeled set, improving the robustness of the segmenter.
\end{itemize}
\subsection{Evaluation metrics}
We evaluate the accuracy of each segmentation method in terms of Dice score \cite{dice1945}, which quantifies the overlap between two anatomical regions. A Dice score of $1$ indicates perfectly overlapping regions, while $0$ indicates no overlap. The predicted segmentation labels are evaluated relative to anatomical labels generated using FreeSurfer~\cite{fischl2012}.
\begin{table}[t]
\small
\centering
\caption{Segmentation performance in terms of Dice score~\cite{dice1945}, evaluated on a held-out test set of $100$ scans. We report the mean Dice score (and standard deviation in parentheses) across all $30$ anatomical labels and $100$ test subjects. We also report the mean pairwise improvement of each method over the SAS baseline.}
\vspace{-10pt}
\begin{tabular}{m{3.3cm}|| M{2cm}|M{2cm}}
\rowgroup{Method}&Dice score&Pairwise Dice improvement\\
\hline\hline
SAS&{0.759 (0.137)}&-\\ 
SAS-aug&{0.775 (0.147)}&0.016 (0.041)\\ 
Rand-aug&{0.765 (0.143)}&{0.006 (0.088)}\\\hline
Ours-coupled &{0.795 (0.133)}&0.036 (0.036)\\
Ours-indep &\textbf{0.804} (0.130)&\textbf{0.045} (0.038)\\
Ours-indep + rand-aug &\textbf{0.815} (0.123)&\textbf{0.056} (0.044)\\\hline
Supervised (upper bound)&{0.849 (0.092)}&{0.089 (0.072)}\\\hline
\end{tabular}
\label{tbl:seg_summary}
\end{table}

\subsection{Results}

\subsubsection{Segmentation performance}
Table \ref{tbl:seg_summary} shows the segmentation accuracy attained by each method. Our methods outperform all baselines in mean Dice score across all $30$ evaluation labels, showing significant improvements over the next best baselines \textit{rand-aug} ($p<1\mathrm{e}\text{-}{15}$ using a paired t-test) and \textit{SAS-aug} ($p<1\mathrm{e}\text{-}{20}$). 

\begin{figure}[t]
        \includegraphics[width=0.95\linewidth]{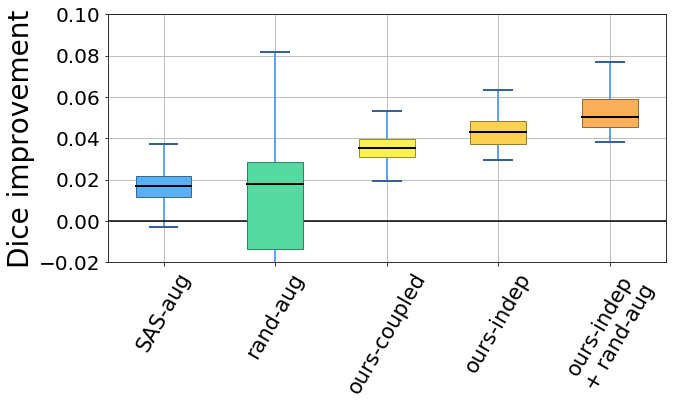}
        \vspace{-8pt}
    \caption{Pairwise improvement in mean Dice score (with the mean computed across all $30$ anatomical labels) compared to the SAS baseline, shown across all test subjects. }\label{fig:dice_vs_sas}
\end{figure}

\begin{figure}[t]
    \centering
    \includegraphics[width=\linewidth]{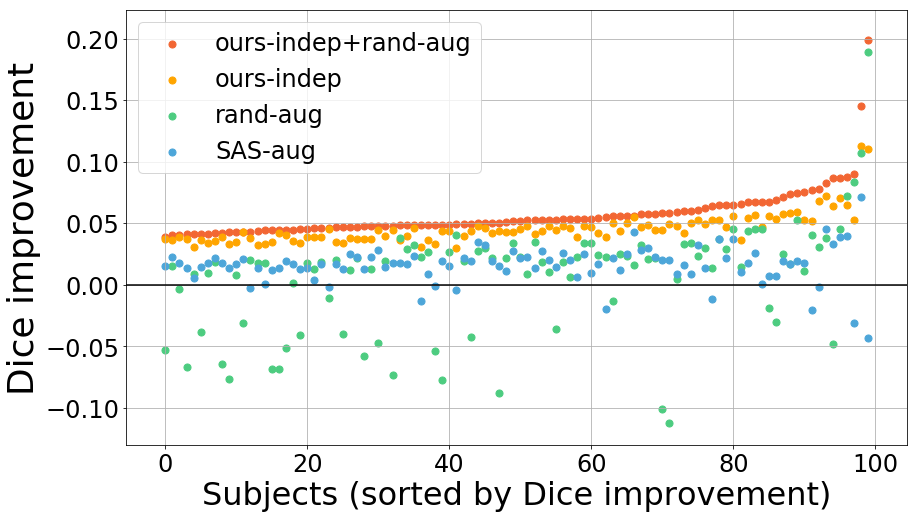}
    \vspace{-14pt}
    \caption{Pairwise improvement in mean Dice score (with the mean computed across all $30$ anatomical labels) compared to the SAS baseline, shown for each test subject. Subjects are sorted by the Dice improvement of \textit{ours-indep+rand-aug} over SAS.}\label{fig:dice_diff}
\end{figure}

In Figs. \ref{fig:dice_vs_sas} and \ref{fig:dice_diff}, we compare each method to the single-atlas segmentation baseline. Fig. \ref{fig:dice_vs_sas} shows that our methods attain the most improvement on average, and are more consistent than hand-tuned random augmentation. Fig. \ref{fig:dice_diff} shows that \textit{ours-indep + rand-aug} is consistently better than each baseline on every test subject. \textit{Ours-indep} alone is always better than \textit{SAS-aug} and \textit{SAS}, and is better than \textit{rand-aug} on $95$ of the $100$ test scans. 

\begin{figure*}[t]
    \centering
    \includegraphics[width=\linewidth]{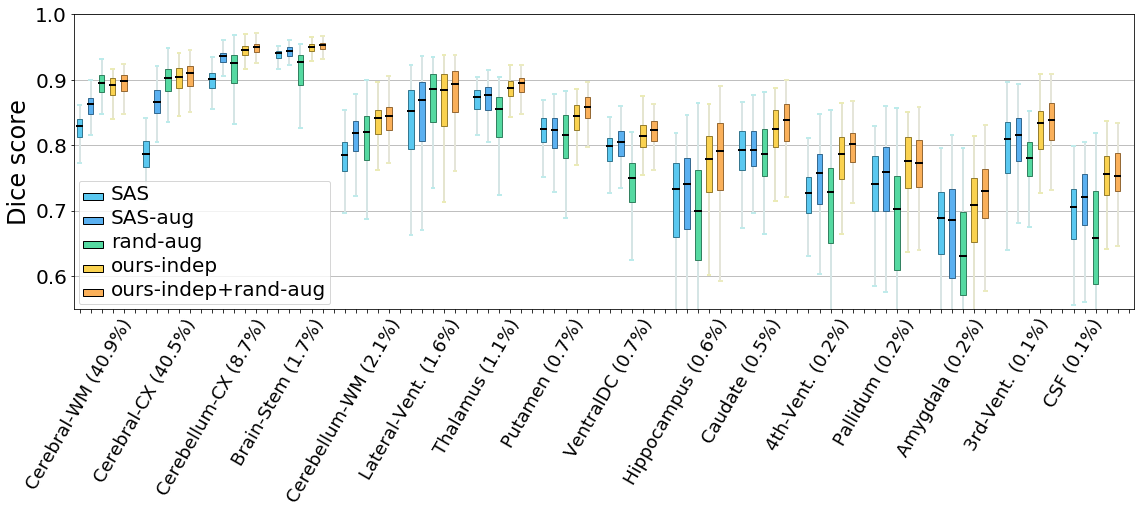}
        \vspace{-12pt}
    \caption{Segmentation accuracy of each method across various brain structures. Labels are sorted by the volume occupied by each structure in the atlas (shown in parentheses), and labels consisting of left and right structures (\textit{e.g.}, Hippocampus) are combined. We abbreviate the labels: white matter (WM), cortex (CX), ventricle (vent), and cerebrospinal fluid (CSF). }\label{fig:dice_per_label}
\end{figure*}

\begin{figure}
    \centering
    \includegraphics[width=\linewidth]{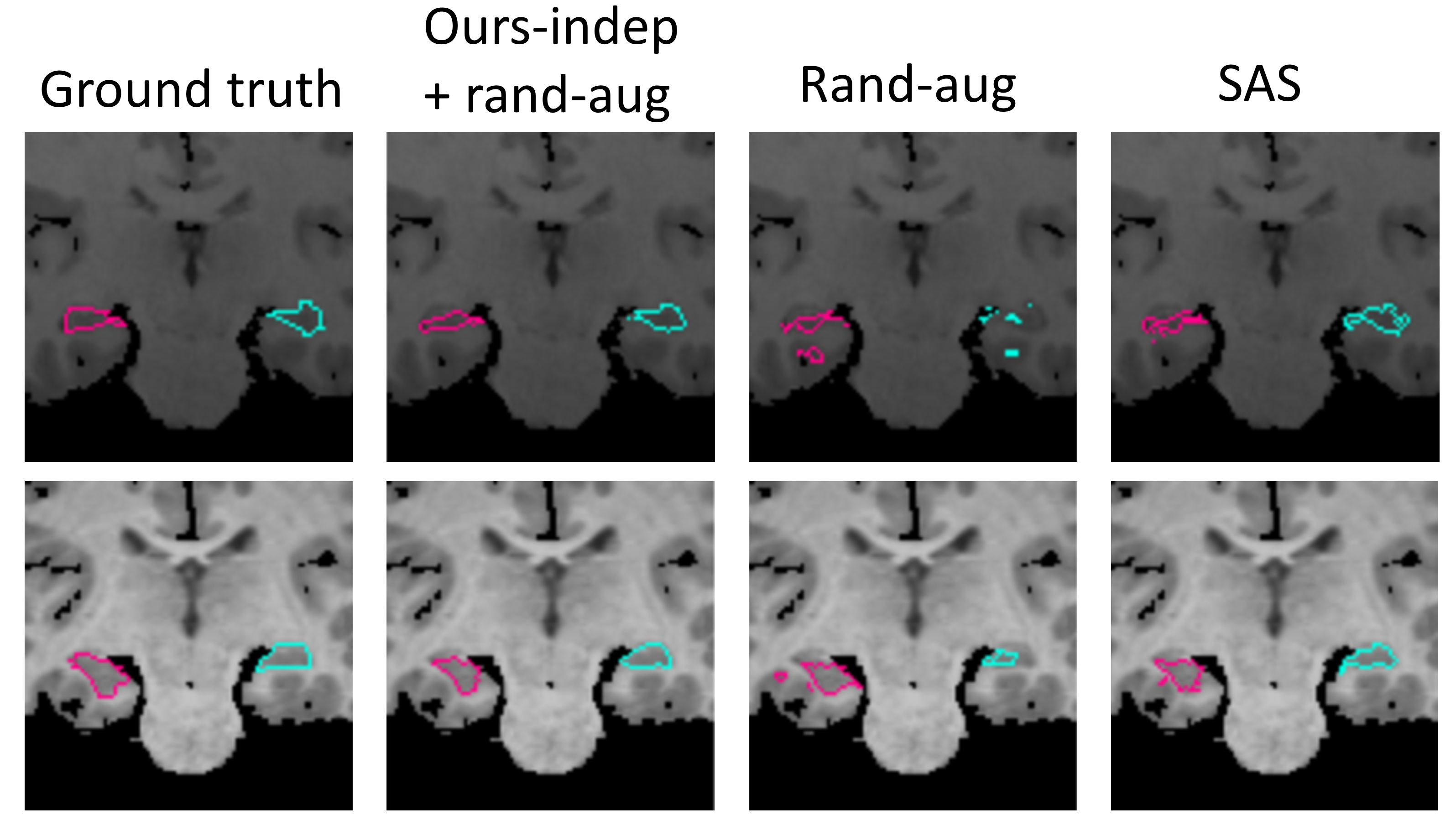}
        \vspace{-5pt}
    \caption{Hippocampus segmentation predictions for two test subjects (rows). Our method (column 2) produces more accurate segmentations than the baselines (columns 3 and 4).}\label{fig:seg_preds}
\end{figure}

Fig.~\ref{fig:dice_per_label} shows that \textit{rand-aug} improves Dice over \textit{SAS} on large anatomical structures, but is detrimental for smaller ones. In contrast, our methods produce consistent improvements over \textit{SAS} and \textit{SAS-aug} across all structures. We show several examples of segmented hippocampi in Fig.~\ref{fig:seg_preds}.


\subsubsection{Synthesized images}
\begin{figure*}[t]
    \centering
        \includegraphics[width=\linewidth]{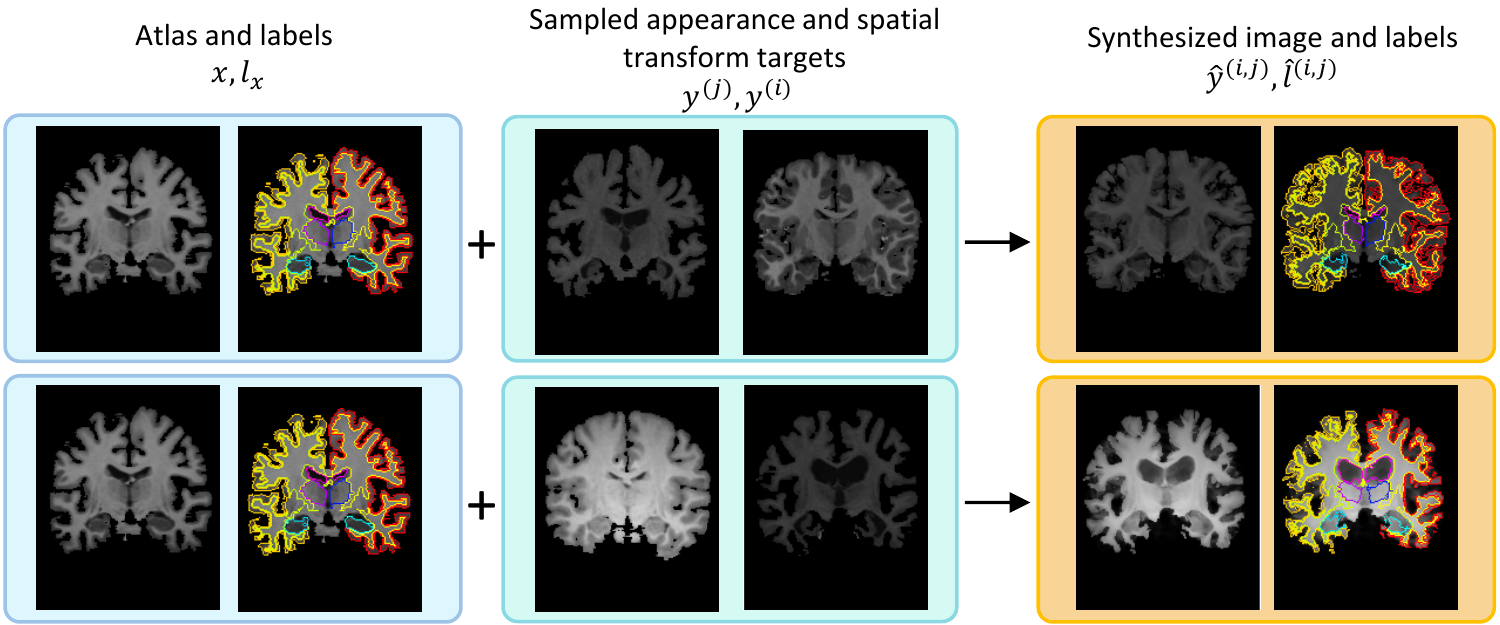}
    \caption{Since we model spatial and appearance transformations independently, we are able to synthesize a variety of combined effects. We show some examples synthesized using transformations learned from the training set; these transformations form the bases of our augmentation model. The top row shows a synthetic image where the appearance transformation produced a darkening effect, and the spatial transformation shrunk the ventricles and widened the whole brain. In the second row, the atlas is brightened and the ventricles are enlarged.}\label{fig:synth}
\end{figure*}


Our independent spatial and appearance models enable the synthesis of a wide variety of brain appearances. Fig. \ref{fig:synth} shows some examples where combining transformations produces realistic results with accurate labels.

\section{Discussion}\label{sec:discussion}
\vspace{-2pt}
\paragraph{Why do we outperform single-atlas segmentation?}
\begin{figure}[t]
\centering
\includegraphics[width=0.8\linewidth]{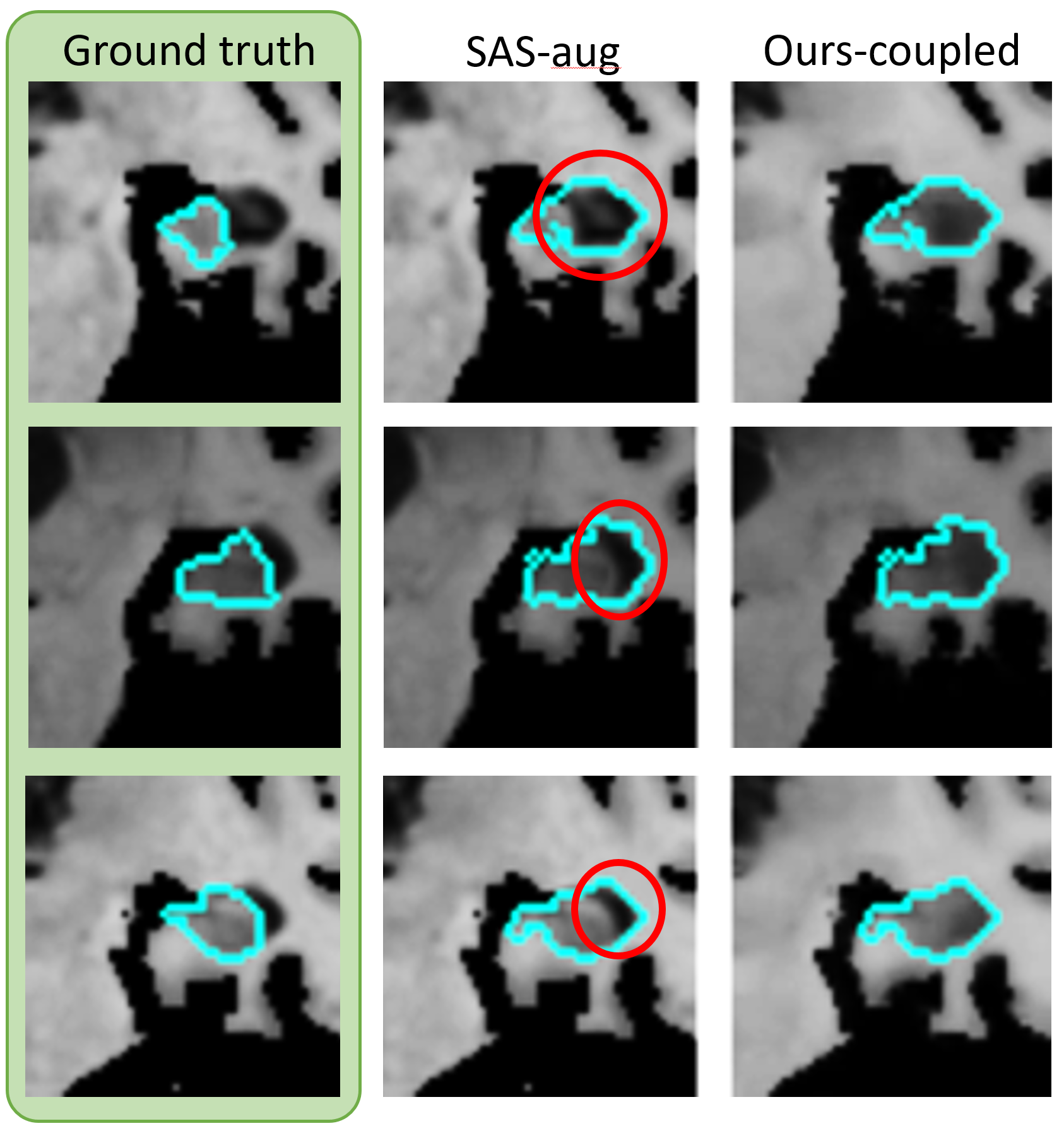}
\caption{Synthetic training examples produced by \textit{SAS-aug} (column 2) and \textit{ours-coupled} (column 3). When the spatial model (used by both methods) produces imperfect warped labels, \textit{SAS-aug} pairs the warped label with incorrect image textures. Our method still produces a useful training example by matching the synthesized image texture to the label.}
\label{fig:us_vs_sasaug}
\end{figure}
Our methods rely on the same spatial registration model that is used for \textit{SAS} and \textit{SAS-aug}. Both \textit{ours-coupled} and \textit{SAS-aug} augment the segmenter training set with $100$ new images. 

To understand why our method produces better segmentations, we examine the augmented images. Our method warps the image in the same way as the labels, ensuring that the warped labels match the transformed image. On the other hand, \textit{SAS-aug} applies the warped labels to the original image, so any errors or noise in the registration results in a mis-labeled new training example for the segmenter. Fig.~\ref{fig:us_vs_sasaug} highlights examples where our method synthesizes image texture within the hippocampus label that is more consistent with the texture of the ground truth hippocampus, resulting in a more useful synthetic training example.\vspace{-8pt}

\paragraph{Extensions} 
Our framework lends itself to several plausible future extensions. In Section \ref{sec:transform_models}, we discussed the use of an approximate inverse deformation function for learning the appearance transformation in the reference frame of the atlas. Rather than learning a separate inverse spatial transform model, in the future we will leverage existing work in diffeomorphic registration \cite{ashburner2007fast,avants2008symmetric,beg2005computing,dalca2018unsupervised,zhang2017frequency}.

We sample transformations from a discrete set of spatial and appearance transformations. This could be extended to span the space of transformations more richly, \textit{e.g.,} through interpolation between transformations, or using compositions of transformations. 

We demonstrated our approach on brain MRIs. Since the method uses no brain- or MRI-specific information, it is feasible to extend it to other anatomy or imaging modalities, such as CT. 

\section{Conclusion}

We presented a learning-based method for data augmentation, and demonstrated it on one-shot medical image segmentation. 

We start with one labeled image and a set of unlabeled examples. Using learning-based registration methods, we model the set of spatial and appearance transformations between the labeled and unlabeled examples. These transformations capture effects such as non-linear deformations and variations in imaging intensity. We synthesize new labeled examples by sampling transformations and applying them to the labeled example, producing a wide variety of realistic new images.

We use these synthesized examples to train a supervised segmentation model. The segmenter out-performs existing one-shot segmentation methods on every example in our test set, approaching the performance of a fully supervised model. This framework enables segmentation in many applications, such as clinical settings where time constraints permit the manual annotation of only a few scans. \\\\

\noindent In summary, this work shows that:
\begin{itemize}
\item learning independent models of spatial and appearance transformations from unlabeled images enables the synthesis of diverse and realistic labeled examples, and

\item these synthesized examples can be used to train a segmentation model that out-performs existing methods in a one-shot scenario.

\end{itemize}

{\small
\bibliographystyle{ieee}
\bibliography{egbib}
}

\end{document}